# Unsupervised Neural Network for Automated Classification of Surgical Urgency Levels in Medical Transcriptions

**Sadaf Tabatabaee and Sarah S. Lam**
*Binghamton University, Binghamton, NY, USA*

## Abstract

Efficient classification of surgical procedures by urgency is paramount to optimize patient care and resource allocation within healthcare systems. This study introduces an unsupervised neural network approach to automatically categorize surgical transcriptions into three urgency levels: immediate, urgent, and elective. Leveraging BioClinicalBERT, a domain-specific language model, surgical transcripts are transformed into high-dimensional embeddings that capture their semantic nuances. These embeddings are subsequently clustered using both K-means and Deep Embedding Clustering (DEC) algorithms, in which DEC demonstrates superior performance in the formation of cohesive and well-separated clusters. To ensure clinical relevance and accuracy, the clustering results undergo validation through the Modified Delphi Method, which involves expert review and refinement. Following validation, a neural network that integrates Bidirectional Long Short-Term Memory (BiLSTM) layers with BioClinicalBERT embeddings is developed for classification tasks. The model is rigorously evaluated using cross-validation and metrics such as accuracy, precision, recall, and F1-score, which achieve robust performance and demonstrate strong generalization capabilities on unseen data. This unsupervised framework not only addresses the challenge of limited labeled data but also provides a scalable and reliable solution for real-time surgical prioritization, which ultimately enhances operational efficiency and patient outcomes in dynamic medical environments.

## 1. Introduction and Related Literature

Clinical notes are crucial patient records, which detail diagnoses, treatments, and care plans not found in structured datasets [1, 2]. Use of NLP and sentiment analysis to examine clinical notes reveals hidden patterns, provides actionable insights, and enhances decision-making [1, 3, 4]. This is vital for tasks such as surgery prioritization, triage, and real-time clinical decision support [2, 5, 6]. Classification of surgical transcripts by urgency is key to enhance patient care and optimize hospital resources [6]. Classification of surgeries as elective, urgent, or immediate helps hospitals streamline scheduling, reduce delays, and avoid resource bottlenecks such as operating room congestion or staff shortages [2, 7, 8]. This underscores the need for research on unsupervised neural networks for automated classification of surgical urgency in medical transcriptions, which is a field that has advanced significantly in healthcare. Starting in 2018, Gencoglu [9] used deep convolutional autoencoders alongside traditional clustering algorithms to significantly improve the clustering of health tweets. By 2021, Alexander et al. [10] identified Alzheimer's disease subtypes using various clustering techniques such as K-means and affinity propagation, which finds K-means most effective. By 2022, Zelina et al. [1] achieved high classification accuracy in segmentation and classification of clinical notes using K-means clustering with a BiLSTM and Transformer model [11]. Most recently in 2023 and 2024, Grissette and Niu [12, 13] reported 87% performance using deep associative neural networks for biomedical sentiment analysis [13]. Frameworks such as National Confidential Enquiry into Patient Outcome and Death (NCEPOD), Trauma and Acute Care Surgery (TACS), and custom hospital protocols categorize surgeries by urgency but often lack scalability, precision, and adaptability across diverse healthcare settings [2, 3, 6, 7]. Recent studies have further emphasized the importance of refining these systems. Iyengar et al. [7] introduced a clinical guide for surgical prioritization (mild, moderate, and severe) during the COVID-19 pandemic to ensure timely intervention.

This research addresses a gap in the literature by using unsupervised sentiment analysis to categorize surgical procedures as *immediate*, *urgent*, or *elective*, based on the NCEPOD framework and existing literature, without relying on explicit labels. The study employs BioBERT, a deep learning model pretrained on biomedical texts, to embed surgery-related transcripts and capture semantic meanings. Unsupervised clustering is used to group the data into urgency levels, which is guided by established frameworks and literature. To address the scarcity of labeled data and improve generalization across clinical datasets, the study incorporates *domain expert validation*, in which healthcare professionals review and refine clustering results to ensure label accuracy for downstream tasks. Building on this, the study aims to develop a robust neural network that integrates BiLSTM layers with BioBERT embeddings for classification, which enhances patient care and resource allocation.

## 2. Methodology

The methodology encompasses three main steps: The first step involves data preprocessing and clustering. Surgical transcripts are thoroughly prepared by advanced clinical note preprocessing steps to improve data quality and consistency. To generate contextualized embeddings of the text, BioClinicalBERT (a specialized variant of the Bidirectional Encoder Representations from Transformers (BERT) architecture pretrained on large-scale biomedical and clinical texts) is used. These embeddings, initially 768-dimensional, are reduced to 50 dimensions using Uniform Manifold Approximation and Projection (UMAP). This dimensionality reduction ensures computational efficiency while preserving the local and global structure of the data. Clustering is then performed using two approaches. The K-means method determines the optimal number of clusters, which is three, based on silhouette analysis. The DEC method, a deep learning-based approach that jointly learns feature representations and cluster assignments, further refines the process by integrating a custom clustering layer with an autoencoder, which enables simultaneous optimization of feature representations and cluster assignments through iterative training. The second step is validation using the Delphi Method. The Modified Delphi Method is employed to validate the clustering results obtained from K-means and DEC. Labels are finalized using a weighted majority approach, which gives priority to DEC outcomes and expert input. This validation ensures that the classification of data is both unbiased and robust. The final step involves the development of a neural network model for classification. This model combines BiLSTM layers with BioBERT embeddings to perform sentiment analysis on the surgical transcripts. The model's effectiveness is assessed through cross-validation across multiple metrics, which include accuracy, precision, recall, and F1-score. It is also tested on unseen data to confirm its ability to generalize across various clinical settings. This phase highlights the potential of the model to enhance clinical decision-making and demonstrates its broader applicability within healthcare systems. This study utilizes a publicly available dataset that comprises medical transcription samples from various specialties, which addresses the significant challenge of accessing medical data due to HIPAA privacy regulations. This dataset is sourced from mtsamples.com.

### 2.1. Data Preprocessing

This study used intensive data preprocessing to prepare surgical transcription data for analysis and modeling. The dataset was filtered to include only entries categorized under the *Surgery* specialty, which ensures relevance. Rows with missing *transcription* fields were removed to maintain data integrity. Normalization steps included converting text to lowercase, removing punctuation and digits using regular expressions, and leveraging SpaCy for tokenization and lemmatization, which standardized words to their base forms. Stop words, both general and medicalspecific, were excluded to focus on informative terms. Medical abbreviations were expanded using a dictionary, and key clinical phrases (e.g., emergency surgery) were merged into single tokens to preserve contextual meaning. The processed tokens were recombined into a *final text* column for use in exploratory analyses and modeling. Each preprocessing step was monitored to ensure accuracy, which resulted in a robust dataset prepared for further investigation. To effectively capture the semantic nuances of surgical transcriptions, *BioClinicalBERT* is employed. *BioClinicalBERT* generates contextualized embeddings that comprehend the meaning of words based on their surrounding context. This deep semantic understanding is crucial to accurately interpret complex surgical transcriptions. Textual data was first tokenized using the BioClinicalBERT tokenizer and then converted into high-dimensional (768-dimensional) vector embeddings. To enhance computational efficiency, embeddings were processed in batches of eight, which extract the [CLS] token representations that encapsulate the entire sequence's information. Given the original high dimensionality, UMAP was applied to reduce the embeddings to 50 dimensions, which preserve both local and global data structures and facilitate more effective clustering and visualization. Notably, the final preprocessed dataset comprises 1,322 transcriptions. This substantial sample size underscores the robustness of the dataset and provides a strong foundation for subsequent analytical and modeling phases.

## 3. Results and Discussion

To address the first goal of the study, two principal clustering methods were applied to analyze surgical transcription data: K-means clustering and DEC. The study employed the K-means algorithm to cluster the embedded representations of surgery transcriptions. To determine the optimal number of clusters, silhouette analysis was performed on cluster counts that range from 2 to 9. The silhouette score, which measures how similar each sample is to its own cluster compared to others, indicated that three clusters ($k = 3$) provided a balanced trade-off between cohesion and separation. The K-means algorithm was then initialized with three clusters and fitted to the UMAP, which reduced embeddings by assigning each transcription to a corresponding cluster for further analysis (see Figure 1).

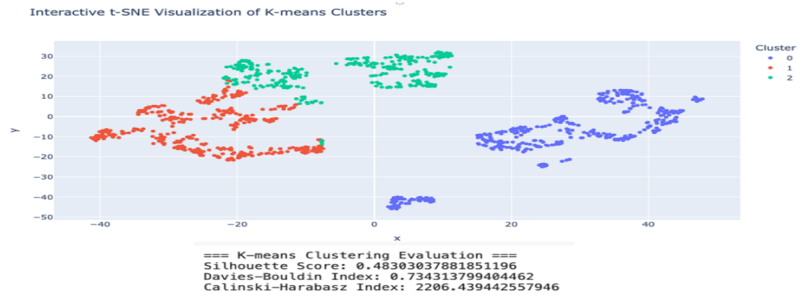

Figure 1: K-means clustering results on surgical transcriptions

The DEC model in this study integrates a deep autoencoder with a specialized clustering layer to enhance clustering performance. The encoder compresses input data with 50 features into a compact 100-dimensional latent space using dense layers with ReLU activations, while the decoder reconstructs the original input, which ensures the encoder captures critical information. A custom clustering layer assigns soft cluster labels based on latent features using a t-distribution. The architecture, which has 652,150 trainable parameters, balances complexity and overfitting risk. It employs Adam for autoencoder training and Stochastic Gradient Descent (SGD) with momentum for clustering refinement, using a composite loss that includes Mean Squared Error (MSE) to measure reconstruction accuracy and Kullback-Leibler Divergence (KLD) to regularize the latent space. The convergence of the DEC model is illustrated in Figure 2. This design allows for rich feature learning, efficient clustering, and scalability, which makes it highly effective for high-dimensional data such as surgical transcriptions. Initial cluster centers were derived from performing K-means on the reduced embeddings. The DEC model was then trained iteratively, which refines both the feature representations and cluster assignments by updating target distributions and monitoring convergence based on changes in cluster labels. Post-training, DEC provided refined cluster labels for comparative analysis with K-means clusters (see Figure 3).

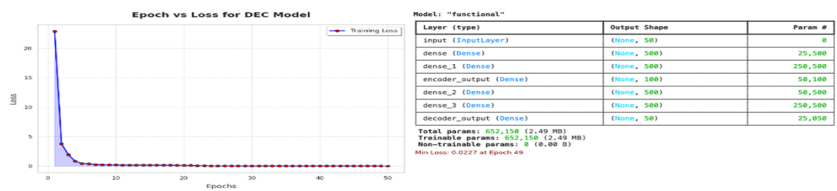

Figure 2: Training dynamics and convergence of the DEC model

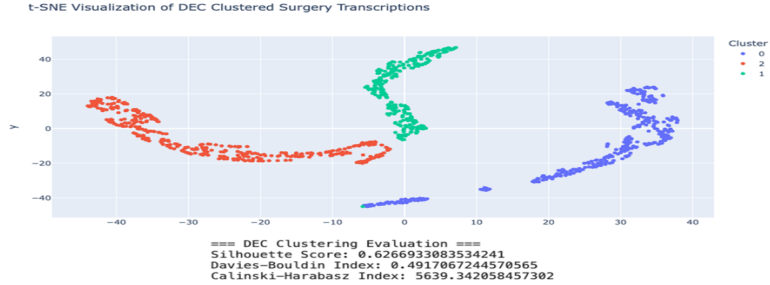

Figure 3: DEC clustering results on surgical transcriptions

The DEC model achieved a higher Silhouette Score of 0.627 compared to K-means' 0.483. This indicates that DEC forms more cohesive and well-separated clusters, which demonstrates its superior effectiveness in the accurate grouping of the surgical transcription data. In summary, the DEC model demonstrates a clear advantage over K-means in clustering surgical transcription data. Its higher Silhouette and Calinski-Harabasz scores, coupled with a lower Davies-Bouldin Index, validate its effectiveness in producing well-defined, distinct clusters. This makes DEC a more suitable choice for complex text-based clustering tasks, where capture of intricate semantic relationships is crucial for meaningful data segmentation.

### 3.1. Validation Using the Delphi Method

Following the identification of unsupervised labels through clustering, the study transitions into the validation phase to ensure the labels' reliability and practical utility. This involves incorporating Domain Expert Validation, in which healthcare professionals review and refine the clustering results generated by K-means and DEC methods to enhance their clinical relevance.

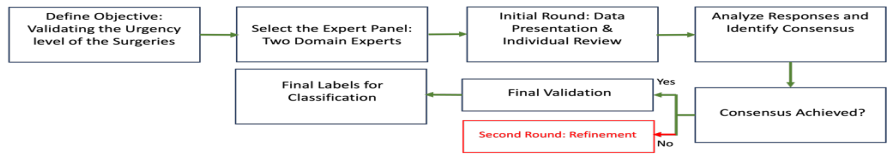

Figure 4: Process of modified Delphi method for urgency level of surgeries' labels validation

Two domain experts participated in the validation process. First, a subset of the data (20% approximately 250 transcripts) was selected for review, in which priority is given to the 10 cases where K-means and DEC clustering methods disagreed. For these cases, the final label was determined using a weighted majority approach that was based on domain expertise and heuristic judgment, which combined inputs from expert assessments (a weight of 0.6), the DEC method (a weight of 0.3), and K-means clustering (a weight of 0.1). The remaining 240 transcripts were balanced across the three categories (elective, urgent, and immediate) and randomly selected within each cluster. The two experts independently reviewed the assigned labels and provided their opinions without knowing each other's feedback, which ensures unbiased evaluations. For the cases where both clustering methods agreed, the result of the Delphi method also showed the agreement, which confirms the reliability of the clustering approach. After all feedback was integrated, the final validated labels were established, creating the ground truth for the subsequent classification task. This iterative and structured approach allowed for a robust validation process that effectively integrated algorithmic clustering with expert domain knowledge. Subsequently, the study focuses on development of a neural network that integrates BiLSTM layers with BioBERT embeddings or employs a transformer-based architecture to perform sentiment analysis on surgical transcriptions.

### 3.2. Neural Network for Classification

The classification task utilizes BioClinicalBERT embeddings, and a custom neural network designed for sequence classification, which leverages advanced preprocessing, stratified splits, and hyperparameter optimization for optimal performance. The steps involve data cleaning, tokenization, and augmentation, followed by training and evaluating the model. Key highlights include the application of stratified splits to balance the dataset across urgency classes, which handles class imbalance using computed class weights, and hyperparameter tuning with Optuna to achieve the best performance. The classification model was trained on the labeled data obtained from the previous steps. The dataset was prepared for the classification task by performing thorough preprocessing and balancing the class distribution. The target labels were mapped as follows: Elective (0), Immediate (1), and Urgent (2), in which three distinct classes were identified for the study. Text data underwent cleaning to remove unnecessary characters, punctuation, and whitespace, and advanced preprocessing included tokenization, lemmatization, and stop word removal, which produced text ready for analysis. After filtering, all classes were retained, with the final distribution as Urgent (408 samples), Elective (388 samples), and Immediate (292 samples). The data was split into an 80% training set (870 samples) and a 20% testing set (218 samples). The training set distribution was Urgent (326 samples), Elective (310 samples), and Immediate (234 samples), while the testing set distribution mirrored the class balance, which ensures a fair evaluation. Text length analysis revealed a mean of 249 words and a maximum of 1,049 words, in which the model input was capped at 512 tokens for compatibility with transformer models such as BioClinicalBERT. The custom neural network integrates BioClinicalBERT embeddings with a classification model tailored to identify the urgency level of surgeries. The network begins with BioClinicalBERT, a pretrained transformer model, to extract domain-specific embeddings from tokenized doctor notes, which encodes rich contextual information for each token. These embeddings are passed through a BiLSTM layer, which captures sequential dependencies in both forward and backward directions, this enhances the understanding of the patient's situation. A dropout layer is incorporated to prevent overfitting by randomly deactivating certain connections during training. The output from the BiLSTM is reduced through a fully connected (dense) layer, which maps the high-dimensional contextualized features to the target labels—Elective, Urgent, or Immediate. Finally, a softmax activation function converts the dense layer's outputs into class probabilities, which provides a clear prediction for each surgery's urgency. This architecture leverages both domain knowledge from BioClinicalBERT and sequential modeling capabilities from BiLSTM, which ensures robust performance for the classification task. The confusion matrix illustrates the classification performance of the model for three classes: Elective, Immediate, and Urgent. Each cell represents the number of predictions for a specific combination of actual and predicted labels. The diagonal cells (Elective:75; Immediate:50; Urgent:76) indicate correct predictions, which demonstrates high accuracy for each class. Off-diagonal cells represent misclassifications, in which most errors occur between Immediate and other classes (e.g., 8 Immediate cases misclassified as Elective and 6 Urgent cases misclassified as Immediate). There are no major misclassifications between Elective and Urgent classes, which indicates the model's ability to distinguish these categories effectively. Overall, the matrix shows strong classification performance, in which minor misclassifications are likely due to overlapping features between similar classes.

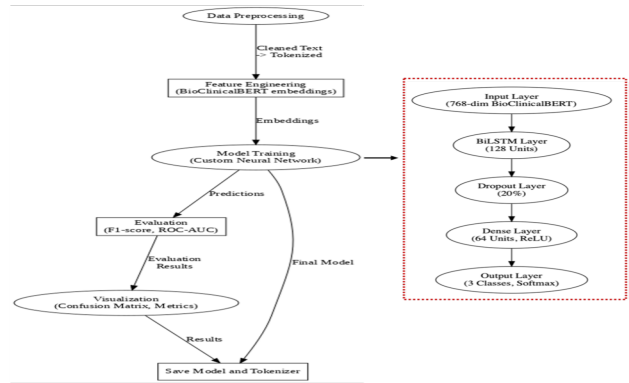

Figure 5: Illustration of the process and neural network architecture used for classification

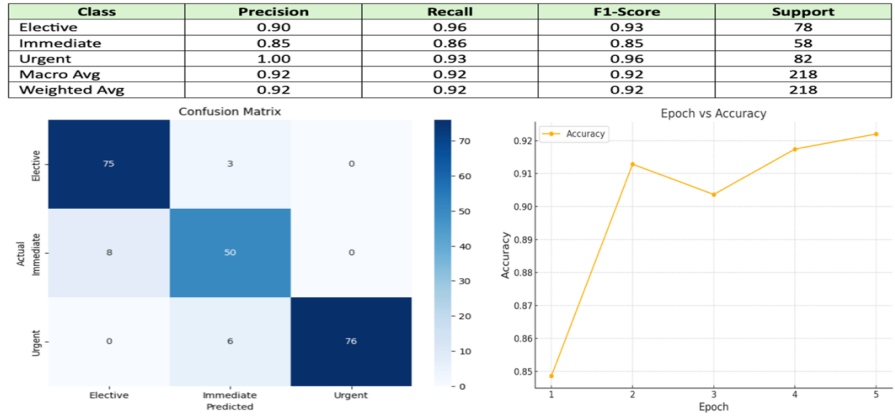

Figure 6: Results of the neural network classification results

## 4. Conclusion

In conclusion, this study demonstrates the viability and effectiveness of employing unsupervised neural network techniques, specifically Deep Embedding Clustering (DEC) combined with advanced NLP models such as BioClinicalBERT, to classify surgical procedures into urgency levels without relying on predefined labels. Compared to traditional K-means clustering, DEC provided more cohesive and well-separated clusters, which effectively captures the nuanced semantic relationships within surgical transcripts. The integration of domain expert validation further ensured that these automatically generated clusters aligned with real-world clinical expertise, which enhances label accuracy and clinical relevance. Building on this robust, unsupervised foundation, the subsequent development of a neural network classification model that incorporates BioClinicalBERT embeddings and BiLSTM layers showcased promising accuracy, precision, and F1-scores. This end-to-end approach—from unsupervised clustering and expert-driven refinement to a fully trained classification model—offers a scalable and generalizable framework to improve surgical prioritization and patient care. By reducing reliance on subjective or inconsistent manual labeling and enabling real-time, data-driven insights, this work contributes to more efficient resource allocation, more timely interventions, and overall better outcomes in dynamic healthcare environments. Moving forward, there are several potential avenues to enhance and expand upon this research. Incorporation of larger, more diverse datasets from multiple healthcare institutions could improve the model's robustness and external validity. Additionally, exploring more advanced transformer architectures and fine-tuning approaches may further refine the clustering and classification performance. Finally, integrating contextual clinical features—such as patient comorbidities, historical data, and resource constraints—could provide a richer, multidimensional framework for predictive analytics, which further supports decision-making processes and patient-centered outcomes in clinical practice.